# ZoomTouch: Multi-User Remote Robot Control in Zoom by DNN-based Gesture Recognition

Ilya Zakharkin  
Skolkovo Institute of Science and Technology (Skoltech)  
Moscow, Russia  
ilya.zakharkin@skoltech.ru

Arman Tsaturyan  
Skolkovo Institute of Science and Technology (Skoltech)  
Moscow, Russia  
arman.tsaturyan@skoltech.ru

Miguel Altamirano Cabrera  
Skolkovo Institute of Science and Technology (Skoltech)  
Moscow, Russia  
miguel.altamirano@skoltech.ru

Jonathan Tirado  
Skolkovo Institute of Science and Technology (Skoltech)  
Moscow, Russia  
jonathan.tirado@skoltech.ru

Dzmitry Tsetserukou  
Skolkovo Institute of Science and Technology (Skoltech)  
Moscow, Russia  
d.tsetserukou@skoltech.ru

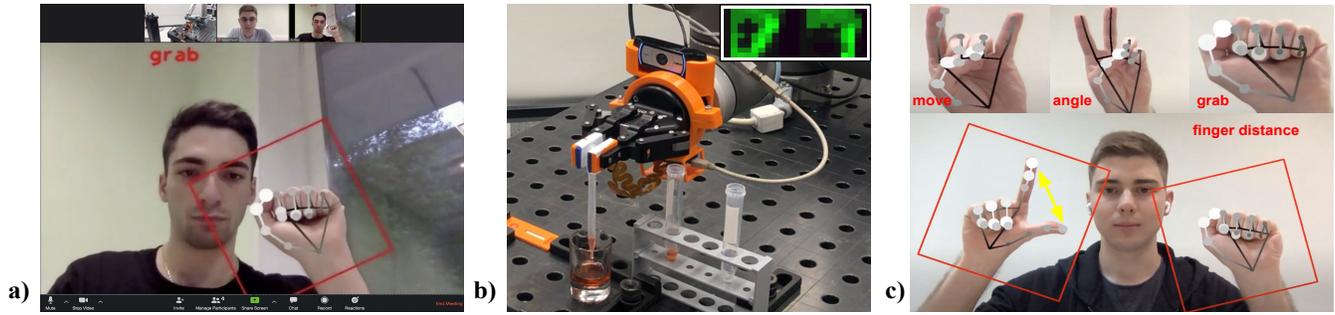

Figure 1: a) Real-time gesture recognition in Zoom by DNN. b) Robot with a gripper that dexterously manipulates the pipettes. c) Available hand gestures for flexible remote control.

## ABSTRACT
We present ZoomTouch - a breakthrough technology for multi-user control of robot from Zoom in real-time by DNN-based gesture recognition. The users from digital world can have a video conferencing and manipulate the robot to make the dexterous manipulations with tangible objects. As the scenario, we proposed the remote COVID-19 test Laboratory to considerably reduce the time to receive the data and substitute medical assistant working in protective gear in close proximity with infected cells. The proposed technology suggests a new type of reality, where multi-users can jointly interact with remote object, e.g. make a new building design, joint cooking in robotic kitchen, etc, and discuss/modify the results at the same time.

## 1 INTRODUCTION

Fully autonomous systems are already essential in industry, science, and medicine. However, an actively developing alternative is a robot-human interaction, where instead of replacing a person with a robot, a person controls the robot [Liu and Wang 2018] and transmit his dexterity. Many methods have been proposed on how to arrange this interaction. Some works use gloves with sensors on hand to control the robotic arm or robot with a gripper [Fang et al. 2015]. To achieve a smooth and safe guiding of a drone formation by a human operator, an impedance control and tactile feedback was proposed in [Tsykunov et al. 2019].

During the COVID-19 pandemic Zoom became an essential tool for education, business meeting, family communication, etc. However, it is not possible to jointly develop a hardware technology in factory environment, or to make a medical tests and scientific experiments remotely.The paper focuses on ZoomTouch to allow several users of Zoom to control the robot located remotely. ZoomTouch consists of Zoom, DNN gesture recognition engine, and remote robot with embedded high fidelity tactile sensor. We conducted experiment on the gesture recognition and real-time robot control to perform the simulation of COVID-19 test. The demonstration revealed a very robust performance of gesture recognition algorithm and agility of robot manipulating soft object, i.e., pipette.

## 2 ZOOMTOUCH PIPELINE

The ZoomTouch system software consists of three parts: screen capture module, hand tracking module, and robot control module. Hardware part is made up of four components: user's laptop, compute station Figure 1(a), a 6 Degrees of Freedom (DoF) collaborative robot UR10 from Universal Robots (b) and a 2-finger gripper from Robotiq endowed with a tactile sensor array which comprises 10 by 10 pressure sensing points [Yem et al. 2019] (c).

When user launches the video conference on the laptop the same conference is opened on the screen of the compute station, this station runs the software to perform screen capture, allowing hand tracking and gesture recognition right from the video conference stream. Eventually, the information is sent to the robot control module that moves the robot itself. Thus ZoomTouch allows a person to control a robot remotely using only laptop webcam and video conference software like Zoom, Google Meet, Microsoft Teams, or Skype in real-time.

The *screen capture module* is a virtual camera that captures the screen using the Open Broadcaster Software (OBS) for video recording and live streaming. The camera stream is then obtained using OpenCV library inside of the hand tracking module.

The *hand tracking module* is implemented on the base of Mediapipe framework. It tracks the landmarks of a hand (or both hands) using which we recognize different hand gestures in real-time on CPU. In this demo we classify 5 gestures: *move*, *angle*, *grab*, *fingerdistance*, and *nogesture* (Figure 1, c)). *Fingerdistance* gesture recognized using the *grab* gesture on the other hand. We trained Gradient Boosting classifier on manually labeled gesture dataset of 1000 images: 200 per *move*, *angle* and *grab* classes, and 400 for *nogesture* class. We used normalized landmarks, angles between joints and pairwise landmark distances as features to predict the gesture class. It resulted in multi-class accuracy of 91% on a test set. The landmarks and gesture on each video frame are passed to the robot control module via the high-performance asynchronous messaging library *ZeroMQ*.

The *robot control module* uses current hand location and gesture information to change the position of the 6DoF UR10 robot TCP position, angle, and the state of the 2-finger gripper (e.g., close, open, or an open distance given by a gesture data).

The gripper also has embedded *two sensors arrays*. Each array is capable of sensing a maximum frame area of 5.8 cm2 with a resolution of 100 points per frame. The sensing frequency is 120 Hz (frames per second). This system is responsible for sensing the pressure applied to instruments or objects grasped by the robot. The pressure data can be used as a security parameter for fragile objects and tactile feedback in teleoperation systems.

## 3 APPLICATIONS

During the demo, the user will seat in front of a laptop in a video-conference with a robot via Zoom. He will see the robot and will be able to control it with gestures. As a small demo, will be possible to move various objects on the table, lift them and drop.

The main demonstration will be the opportunity to control robot for remote medical test. The user will have to guide the robot to take the pipette from the table, put it in a test tube with blood (red color liquid), take a blood sample, and then bring it to the antibodies test and drop blood into the blood compartment.

Our system can be used in many other teleoperation scenarios, e.g. for making experiment on International Space Stations or spectrometer experiment of MARS 2020 space mission to get the insight of Martian soil composition. Multi-user functionality of the system is addressing an actual problem of effective remote collaboration, allowing to work on projects which need manual remote control in a group of several people. It is worth to mention that ZoomTouch could also be used for other types of hardware form factors, such as edge devices, mobile robots and drones.

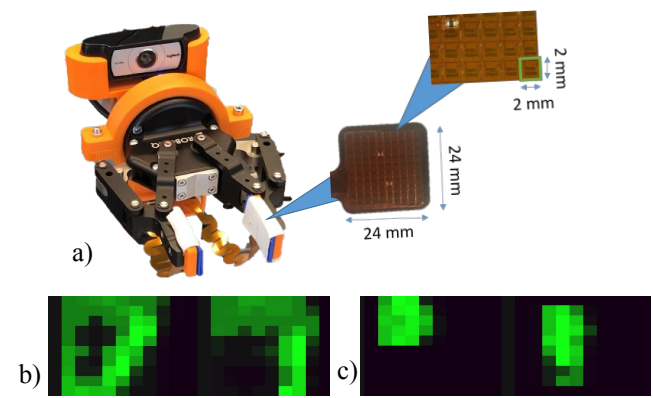

Figure 2: High fidelity tactile sensor array: a) Array placement on the gripper. b) Sensor data when the gripper takes a pipe. 3) Sensor data when the gripper takes a tube.